\documentclass[11pt,a4paper]{article}
\usepackage{coling2020}
\usepackage{times}
\usepackage{latexsym}

\usepackage{floatrow}
\newfloatcommand{capbtabbox}{table}[][\FBwidth]
\usepackage{sidecap} 
\usepackage{caption}
\newsavebox{\figurebox}
\usepackage{hyperref}
\usepackage{url}
\colingfinalcopy
\usepackage{nopageno} 

\usepackage{graphicx} 
\usepackage{multicol}
\usepackage{booktabs} 
\usepackage{algorithm} 
\usepackage{algorithmic} 
\usepackage{color}
\usepackage{amsmath}
\usepackage{amssymb}
\usepackage{amsfonts}
\usepackage{multirow, varwidth}
\usepackage{stfloats} 
\usepackage{verbatim}
\usepackage{enumitem}
\usepackage{subcaption}
\makeatletter
\newif\if@restonecol
\makeatother
\usepackage{xr}
\usepackage[amsmath,thmmarks]{ntheorem}

\usepackage{xspace}
\makeatletter
\DeclareRobustCommand\onedot{\futurelet\@let@token\@onedot}
\def\@onedot{\ifx\@let@token.\else.\null\fi\xspace}

\def\eg{\emph{e.g}\onedot} 
\def\ie{\emph{i.e}\onedot}

\def\etc{\emph{etc}\onedot}

\makeatother

\usepackage[colorinlistoftodos]{todonotes}
\newcommand{\agf}[1]{\vphantom{#1}}  

\newcommand{\simmc}{SIMMC\xspace}
\newcommand{\furniture}{SIMMC-Furniture\xspace}
\newcommand{\fashion}{SIMMC-Fashion\xspace}
\newcommand{\furniturefull}{SIMMC-Furniture (VR)\xspace}
\newcommand{\fashionfull}{SIMMC-Fashion (Image)\xspace}
\newcommand{\platform}{SIMMC Platform\xspace}

\newcommand{\activity}{activity\xspace}
\newcommand{\Activity}{Activity\xspace}
\newcommand{\activities}{activities\xspace}
\newcommand{\Activities}{Activities\xspace}

\newcommand{\reffig}[1]{Fig.~\ref{#1}}
\newcommand{\refsec}[1]{Sec.~\ref{#1}}
\newcommand{\reftab}[1]{Tab.~\ref{#1}}
\newcommand{\refeq}[1]{Eq.~\ref{#1}}

\newcommand{\reportvalbrief}[2]{#1}

\usepackage{booktabs,arydshln}
\makeatletter
\def\adl@drawiv#1#2#3{%
        \hskip.5\tabcolsep
        \xleaders#3{#2.5\@tempdimb #1{1}#2.5\@tempdimb}%
                #2\z@ plus1fil minus1fil\relax
        \hskip.5\tabcolsep}
\newcommand{\cdashlinelr}[1]{%
  \noalign{\vskip\aboverulesep
           \global\let\@dashdrawstore\adl@draw
           \global\let\adl@draw\adl@drawiv}
  \cdashline{#1}
  \noalign{\global\let\adl@draw\@dashdrawstore
           \vskip\belowrulesep}}
\makeatother

\belowdisplayskip \abovedisplayskip

\newlength{\sectionReduceTop}
\newlength{\sectionReduceBot}
\newlength{\subsectionReduceTop}
\newlength{\subsectionReduceBot}
\newlength{\abstractReduceTop}
\newlength{\abstractReduceBot}
\newlength{\captionReduceTop}
\newlength{\captionReduceBot}
\newlength{\subsubsectionReduceTop}
\newlength{\subsubsectionReduceBot}

\newlength{\eqnReduceTop}
\newlength{\eqnReduceBot}

\newlength{\horSkip}
\newlength{\verSkip}

\newlength{\figureHeight}
\usepackage{marvosym}
\usepackage[dash,dot]{dashundergaps}
\usepackage{wrapfig}

\newcommand{\sk}[1]{{\color{black}{#1}}} 

\newcommand*\samethanks[1][\value{footnote}]{\footnotemark[#1]}

\title{Situated and Interactive Multimodal Conversations\vspace{-.2in}}

\author{\\
\textbf{Seungwhan Moon}\thanks{\hspace{10pt}Joint first authors.\hspace*{17pt}\Letter\hspace{2pt}\{shanemoon, skottur\}@fb.com.},
\textbf{Satwik Kottur}\samethanks,
\textbf{Paul A. Crook}\thanks{\hspace{10pt}Joint second authors.\hspace*{6pt}\Letter\hspace{2pt}\{pacrook, deankita, shivanip\}@fb.com.},
\textbf{Ankita De}\samethanks,
\textbf{Shivani Poddar}\samethanks\\
\textbf{Theodore Levin},
\textbf{David Whitney},
\textbf{Daniel Difranco},
\textbf{Ahmad Beirami} \\
\textbf{Eunjoon Cho},
\textbf{Rajen Subba},
\textbf{Alborz Geramifard} \\~\\
{Facebook} \\
{\Letter\hspace{3pt}\texttt{simmc@fb.com}}\vspace{-.1in}
}
\date{}

\begin{document}
\maketitle
\thispagestyle{plain}
\pagestyle{plain}

\begin{abstract}
\vspace{-5pt}
Next generation virtual assistants are envisioned to handle 
multimodal inputs (\eg, vision, memories of previous interactions, and the user's utterances), and  perform
multimodal actions (\eg, displaying a route while generating the system's utterance).
We introduce  Situated Interactive MultiModal Conversations (\simmc) as a new direction aimed at training agents that take multimodal actions grounded in a \emph{co-evolving} multimodal input
context in addition to the dialog history.
We provide two SIMMC datasets totalling $\sim$13K
human-human dialogs ($\sim$169K utterances) collected using a multimodal Wizard-of-Oz
(WoZ) setup, on two shopping domains:
(a) furniture -- grounded in a shared virtual environment; and
(b) fashion -- grounded in an evolving set of images.
Datasets include multimodal context of the items appearing in each scene, and contextual NLU, NLG and 
coreference annotations using a novel and unified framework of SIMMC \emph{conversational acts} for both user and assistant utterances.

Finally, we present several tasks within \simmc as objective evaluation protocols,
such as structural API prediction, response generation, and dialog state tracking.
We benchmark a collection of existing models on these \simmc tasks as
strong baselines, and demonstrate rich multimodal conversational interactions.
Our data, annotations, and models are publicly
available.\footnote{\url{https://github.com/facebookresearch/simmc}} 
\end{abstract}


\blfootnote{
    This work is licensed under a Creative Commons 
    Attribution 4.0 International License.
    
    \hspace{0.21cm} License details: \url{http://creativecommons.org/licenses/by/4.0/}
}

\vspace{-5pt}
\section{Introduction}
\vspace{-5pt}
As virtual digital assistants become increasingly ubiquitous, they are expected to be embedded in the day-to-day life of users the same way a human assistant would.
We thus envision that the next generation of virtual assistants will be able to process multimodal inputs and provide multimodal outputs beyond the traditional NLP stack.
%
To this end, we present {\bf Situated Interactive MultiModal Conversations (\simmc)}
tasks and datasets as a starting point
in this new research direction.
Specifically, SIMMC focuses on \textbf{task-oriented} dialogs that encompass a \textbf{situated multimodal context},
where situated implies that the user and assistant are continually co-observing the same context, and that context can be updated on each turn.
We provide two new \simmc datasets in the domain of interactive shopping, collected using the
\platform~\cite{simmc}: (1) Furniture and (2) Fashion.
Moreover, we provide fine-grained annotations to allow for both
end-to-end and component-level modelling. The annotation includes natural language understanding (NLU), multimodal-coreference, multimodal state tracking, assistant actions,
natural language generation (NLG), and item appearance logs.

 \begin{figure}[t!]
  \centering
  \includegraphics[width=1.0\columnwidth]{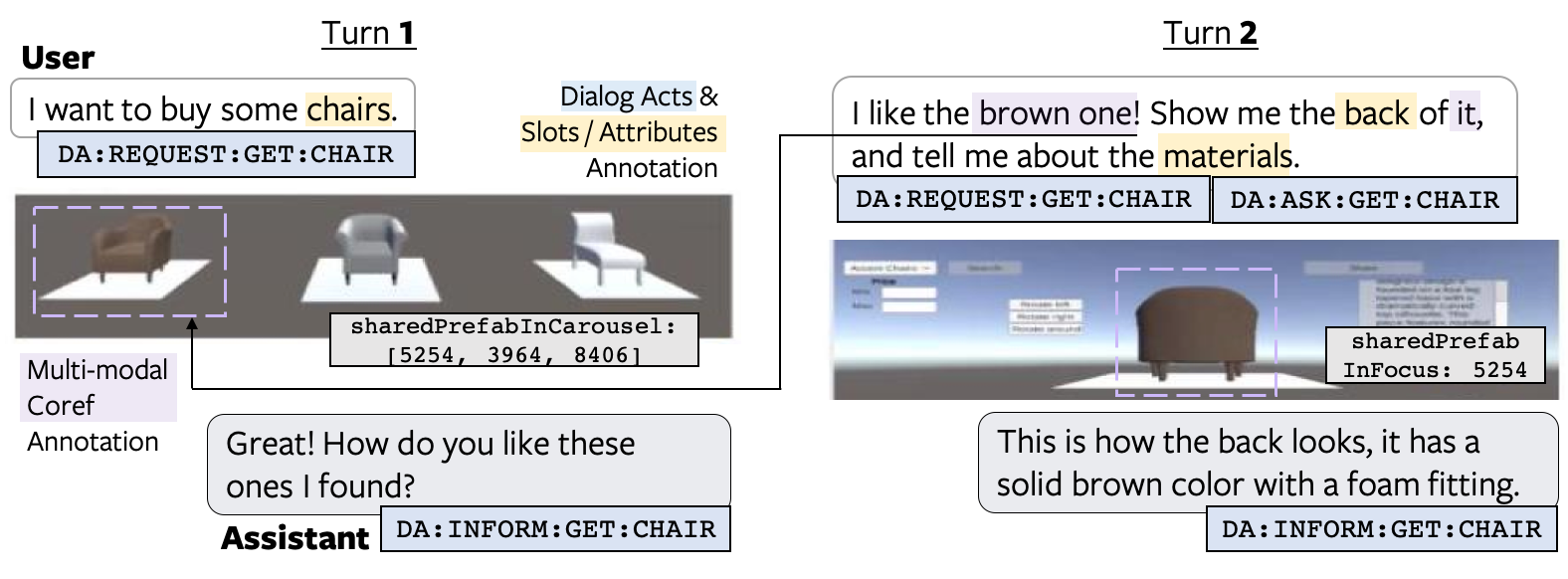}
  \caption{Illustration of a \simmc dialog: a user and an assistant interact in a co-observed multimodal environment for a shopping scenario. The dialog is grounded in an \textit{evolving} multimodal context. The ground-truth of which items (\eg, prefabs) appear is known for each view.} 
  \label{fig:teaser}
\end{figure}

\reffig{fig:teaser} illustrates an exemplary dialog from our \furniture dataset,
where a user is interacting with an assistant with the goal of browsing 
for furniture.
In our setting, the assistant can update the co-observed environment to create a new multimodal context based on the preceding dialog with the user, \eg, visually
presenting recommended chairs in a virtual reality (VR) environment, or responding to the request ``I like the \underline{brown one}. \textit{Show me the back} of \underline{it}." by executing the actions of \textit{focusing on}, and \textit{rotating} the indicated \underline{item}.
These actions change the shared multimodal context, which grounds the next part of the dialog.
The example also highlights novel challenges such as multimodal action prediction (\textit{italics} above) and multimodal coreference resolution (\underline{underlined} elements above).

\section{Novelty \& Related Work}
\label{sec:related_work}

\reftab{tab:related_work} presents main distinctions of \simmc compared to the the existing multimodal dialog datasets.

\noindent \textbf{Multimodal datasets for co-observed, real-world assistant}.
With the ultimate goal of laying the foundations for the real-world
assistant scenarios, we assume a co-observed multimodal context between
a user and an assistant. 
This shifts the primary focus onto the core problem of grounding
conversations in the co-observed multimodal context.
In contrast, the existing literature \cite{visdial,clevr-dialog,guesswhat,talk-the-walk}, drawing motivation from the Visual Question Answering \cite{vqa}, posits the roles of a primary and secondary observer, \ie, \textit{questioner} and \textit{answerer}, who do not co-observe the same
multimodal context.

In addition, we study scenarios in which the situated multimodal context is dynamically updated, reflecting the agent actions.
In our settings, agent actions can be enacted on both the object-level -- changing the view of a specific object within a scene, and the scene-level -- introducing a new scene or an image.
While the dialog-based image retrieval tasks \cite{dialog-image-retrieval,mmd} and the visual navigation tasks 
\cite{vision-dialog-navigation,talk-the-walk}
do comprise context updates, they are limited to the introduction of new visual scenes, \eg, new images or locations.


\noindent \textbf{Focus on task-oriented dialogs}.
We frame the problem as a \textit{task-oriented}, multimodal dialog system, with the aim of extending the capabilities of digital assistants to real-world multimodal settings.
While one focus area of the dialog community is on \textit{task-oriented} dialog, which has practical applicability consumer-facing virtual assistants \cite{dstc2,multiwoz,multiwoz2.1,sgd-dst,chen-etal-2020-jddc}, 
this form of dialog is often neglected in many existing multimodal `dialog' datasets (both in terms of the task design and the annotations), where the primary focus lies in visual grounding of language.
Our work aims to bring important challenges actively studied in the dialog community to the multimodal setting. 
Specifically, we study the multimodal extension of the traditional dialog state tracking (DST) and the assistant API prediction tasks, which have been the key focus of dialog literature \cite{trade,bert-dst-alexa,bert-dst-cmu}.

Compared to the conventional task-oriented dialog datasets (\eg, MultiWoZ \cite{multiwoz}), the agent actions in \simmc span across a
diverse multimodal action space (\eg, \textsc{rotate}, \textsc{search}, \textsc{add\_to\_cart}).
Our study thus shifts the focus of the visual dialog research from the token-level grounding of visual scenes to the task-level understanding of dialogs given multimodal context.


\begin{table*}[t]
    \begin{center}
    \scalebox{0.65}{
    \setlength\tabcolsep{7pt}
    \begin{tabular}{cccccccc}
    \toprule[\heavyrulewidth]
    \multirow{2}{*}{\textbf{Dataset}} & \multirow{2}{*}{\textbf{Modality}} & \multirow{2}{*}{\textbf{Task}} & \multicolumn{2}{c}{\textbf{Provided Context}} & \textbf{Updated} & \textbf{Annotation} & \multirow{2}{*}{ } \\
    \cmidrule(r){4-5}
    & & & \textbf{Q'er} & \textbf{A'er} & \textbf{Context} & \textbf{Granularity} & \\
    \midrule
    \midrule
    Visual Dialog \cite{visdial} & Image & Q\&A & N/A & Visual & N/A & N/A &  \\
    CLEVR-Dialog \cite{clevr-dialog} & Simulated & Q\&A & N/A & Visual & N/A & N/A &  \\
    GuessWhat \cite{guesswhat} & Image & Q\&A & N/A & Visual &  N/A &  N/A &  \\
    Audio Visual Scene-Aware Dialog \cite{avsd-dstc8} & Video & Q\&A & N/A & Visual &  N/A &  N/A &  \\
    TalkTheWalk \cite{talk-the-walk} & Image & Navigation & Visual & Visual + Meta & Location & U $\leftrightarrow$ A &  \\
    Visual-Dialog Navigation \cite{vision-dialog-navigation} & Simulated & Navigation & Visual & Visual + Meta & Location & U $\leftrightarrow$ A &  \\
    Relative Captioning \cite{dialog-image-retrieval} & Image & Image Retrieval & Visual & Visual + Meta & New Image & U $\leftrightarrow$ A &  \\        
    MMD \cite{mmd} & Image & Image Retrieval & Visual & Visual + Meta & New Image & U $\leftrightarrow$ A &  \\        
    \midrule
    \textbf{SIMMC (proposed)} &  \textbf{Image/VR} & \textbf{Task-oriented} & \textbf{Visual} & \textbf{Visual + Meta} & \textbf{Situated} & \textbf{U $\leftrightarrow$ A + Semantic} &  \\
    \bottomrule[\heavyrulewidth]
    \end{tabular}
    }
\end{center}

    \caption{\textbf{Comparison with the existing multimodal dialog corpora}. \textbf{Notations}: (U $\leftrightarrow$ A) Utterance to action pair labels. (Task-oriented) Includes API action prediction, Q\&A, recommendation, item / image retrieval and interaction. (Semantic) Dialog annotations such as NLU, NLG, DST, and Coref. (Situated) VR environment and/or new highlighted images. }
    \vspace*{\captionReduceBot}
    \label{tab:related_work}
\end{table*}

\noindent \textbf{Semantic annotations for multimodal dialogs}. Finally, we present a novel flexible schema for \textit{semantic
annotations} that we developed specifically for the natural multimodal
conversations.
The proposed \simmc annotation schema allows for a more systematic and
structural approach for visual grounding of conversations, which is essential
for solving this challenging problem in the real-world scenarios.
To the best of our knowledge, our dataset is the \textit{first} among the related multimodal dialog corpora to provide fine-grained semantic annotations.

\section{\simmc Datasets}
\label{sec:dataset}


\begin{wraptable}{r}{0.42\textwidth}
    \begin{center}
    \vspace{-20pt}
        \scalebox{0.76}{
        \begin{tabular}{lccc}
        \toprule[\heavyrulewidth]
        \multicolumn{1}{c}{\multirow{2}{*}{\textbf{Statistics}}} & \multicolumn{2}{c}{\textbf{Furniture (VR)}} &
        \multirow{2}{*}{\textbf{\shortstack{Fashion\\(Image)}}} \\
        \cmidrule(r){2-3}
          &  Text & Audio$^\dagger$ & \\
        \midrule
        Total \# dialogs & 6.4k & 1.3k & 6.6k \\
        Total \# utterances & 97.6k & 15.8k & 71.2k \\
        Avg \# rounds / dialog & 7.62 & 7.16 & 5.39 \\
        Avg \# tokens (user) & 11.0 & N/A & 11.10 \\
        Avg \# tokens (assistant) & 12.2 & N/A & 10.87 \\
        \bottomrule[\heavyrulewidth]
        \end{tabular}
    }
    \end{center}
    \captionof{table}{\textbf{SIMMC Datasets Statistics}.
    $^\dagger$We also collected additional dialogs in aural medium
    where annotators exchanged audio messages instead of text.}
    \vspace*{\captionReduceBot}
    \label{tab:datasets_statistics}
\end{wraptable}



For \simmc datasets, we focused on the shopping domain as it often induces rich multimodal interactions around browsing visually grounded items.
As shown in \reffig{fig:teaser}, the setup consists of two human workers, \sk{a user
and an assistant}, conversing around a shopping scenario.
\sk{The goal of the user is to interactively browse through an inventory of items 
while that of the assistant is to facilitate this conversation.}
In addition to having an interactive dialog, the assistant manipulates the
co-observed environment to show off items from the shopping inventory.
A conversational assistant model for the \simmc datasets would need to 
(i) understand the user's utterance using both the dialog history and the
state of the environment -- the latter provided as multimodal context, 
and (ii) produce a multimodal response to the user utterance, including
updates to the co-observed environment to convey meaningful information as
part of the user’s shopping experience.
%
%
We provide two \simmc datasets with slightly different setups and modalities. See \reftab{tab:datasets_statistics} for overall statistics.


\label{subsec:dataset:simmc_furniture}
The {\bf \furniturefull Dataset} captures a scenario where a user is interacting with an assistant whilst browsing for furniture, \eg, couch, or side table. Grounded in a VR environment \cite{unitygameengine} the assistant can manipulate items in the scene while engaging in conversation.
We seed the conversation by presenting the user with either a high-level directive such as
\textit{`Shop for a table'} or an image of a furniture item to shop for.
The user is then connected randomly with a human assistant.
The assistant can filter the catalog by attributes such as furniture category, price, color, and material, navigate through the filtered results and share their view with the user.
As part of the dialog, the user can request to look closer at one of the options,
or see other options.
In response, the assistant can either zoom into an item, present an 
alternate view by rotating it, or look at the catalog description to answer further
questions.
To enable this, the environment is designed to transition between 
two states:
(a) \textit{Carousel}, which displays three filtered furniture items 
(top view, \reffig{fig:teaser}); and
(b) \textit{Focused}, which provides a zoomed in view of one item from the \textit{carousel} view
(bottom view, \reffig{fig:teaser}).
The conversation continues for
6--12 turns until the user considers that they have reached a successful outcome.
\agf{How do we define success?}
\reftab{tab:simmc_furniture_ex_1} shows example dialogs. 

\label{subsec:dataset:simmc_fashion}
The {\bf \fashionfull Dataset} represents user interactions with an assistant 
to obtain recommendations for clothing items, \eg, jacket, dress. Conversations are grounded in 
real-world images that simulate a shopping scene from a user’s point-of-view (POV).
At the start of each dialog the user is presented with a randomly selected `seed' image from the catalog to emulate (visually) that they are in the middle of shopping, as well as a sequence of synthetic {\it memories} of `previously viewed items'.
In addition to the user's context, the assistant has access to a broader catalog that allows for information lookup and item recommendation.
We ask the user to browse and explore options by asking the
assistant for recommendations based on the shared attributes, preferences, as
referred from visual scenes, memories, and assistant-recommend items.
The conversation continues for 6--10 turns until the user is assumed to be given a successful recommendation.
Please refer to \reftab{tab:simmc_fashion_ex_1} for example dialogs. 


\label{sec:item_scene_data}

For both datasets, the ground-truth of which items appear in each view is logged and included in the multimodal context. This allows the problem of computer vision to be sidestepped and focus on semantically combining the modalities.
The datasets were collected through the \platform~\cite{simmc}, an
extension to ParlAI \cite{parlai} for multimodal conversational data
collection and system evaluation.
\sk{Note that even though we focus on English in this work, our data collection framework is
language-agnostic and can be easily extended to other languages.}

\begin{figure*}[hb]
    \centering
    \begin{subfigure}[b]{0.23\textwidth}
        \includegraphics[width=\textwidth]{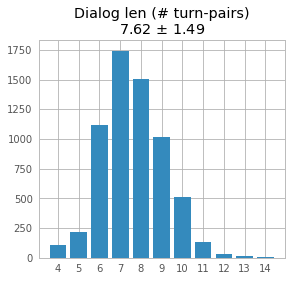}
        \caption{Distribution of Rounds (\furniture)}
        \label{subfig:furniture_rounds_distr}
    \end{subfigure}
    ~\begin{subfigure}[b]{0.24\textwidth}
        \includegraphics[width=\textwidth]{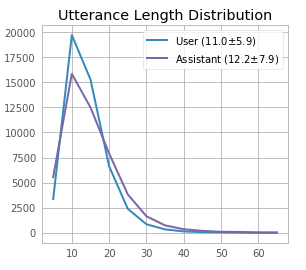}
        \caption{Distribution of Utterance Lens (\furniture)}
        \label{subfig:furniture_utterance_distr}
    \end{subfigure}
    ~\begin{subfigure}[b]{0.23\textwidth}
        \includegraphics[width=\textwidth]{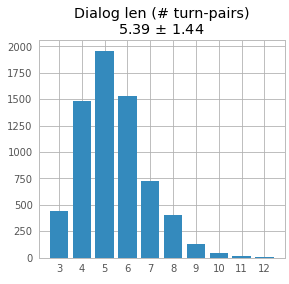}
        \caption{Distribution of Rounds (\fashion)}
        \label{subfig:fashion_rounds_distr}
    \end{subfigure}
    ~\begin{subfigure}[b]{0.24\textwidth}
        \includegraphics[width=\textwidth]{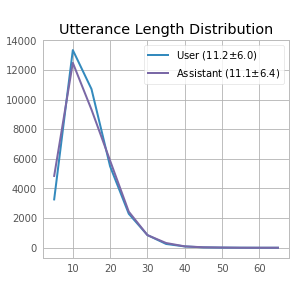}
        \caption{Distribution of Utterance Lens (\fashion)}
        \label{subfig:fashion_utterance_distr}
    \end{subfigure}
    \caption{\textbf{\simmc Datasets Analysis}. Distribution of Rounds and Utterance Lengths (\# of tokens).}
    \label{fig:dataset_analysis}
\end{figure*}

\paragraph{Dataset Analysis.}
\furniture has $6.4k$ dialogs with an average of $7.62$ rounds (or turn pairs)
leading to a total of about $97.6k$ utterances.
Similarly, \fashion consists of $6.6k$ dialogs, each around $5.42$ rounds
 on average, totaling $71.2k$ utterances.
In addition to these sets, we also collect a smaller, audio-based \furniture 
dataset ($1.3k$ dialogs) where the dialog exchanges are aural 
as opposed to written text.

In \reffig{fig:dataset_analysis}, we visualize:
(a) \textit{Distribution of rounds}.
Dialogs in \furniture range from $4$ 
(shorter ones are omitted from the dataset) to a maximum of $14$ rounds, 
with $68\%$ of the dialogs containing $7$--$9$ rounds
(\reffig{subfig:furniture_rounds_distr}).
Dialogs in \fashion range from $3$--$12$ rounds at an average of $5.4 \pm 1.4$ rounds per dialog, as shown in 
\reffig{subfig:fashion_rounds_distr}.
We hope that this widespread range will help train models
 that can handle diverse conversations of varied lengths.
\linebreak
(b) \textit{Distribution of utterance lengths.}
For both user and assistant, we tokenize their utterances and plot the 
distribution in \reffig{subfig:furniture_utterance_distr}.
For \furniture, the assistant utterances are slightly longer with higher
variance at $12.2 \pm 7.9$ when compared to those from the user, at 
$11.0 \pm 5.9$.
A potential reason is that because the assistant has access to the catalog, it is expected
to be more verbose while responding to description related queries (\textit{`User: Tell me more about the brown table'}).
However, we do not observe a similar trend for \fashion where user and assistant turns average around $11.2 \pm 6.0$ tokens per utterance 
(\reffig{subfig:fashion_utterance_distr}).
\linebreak
(c) \textit{Catalog coverage.}
Recall that both \simmc datasets contain conversations in a shopping scenario
grounded in a catalog of furniture and fashion items respectively.
\furniture builds on a catalog of $179$ items, where each dialog contains around 
$3.3$ shares of different views between the user and assistant, and each furniture 
item is shared in roughly $45$ dialogs.
Similarly, \fashion contains $2098$ items that appear in $32$ dialogs on average,
thus providing a rich catalog context to support interesting multimodal dialogs.

\section{\simmc Dialog Annotations}
\label{sec:annotations}
Building a task-oriented multimodal conversational model introduces many new challenges, as it requires both action and item-level understanding of multimodal interactions.
While most of the previous multimodal datasets
provide surface-level annotations 
(\eg, utterance to multimodal action pairs), we believe it is critical to provide the
semantic-level fine-grained annotations that ground the visual context, allowing for a
more systematic and structural study for visual grounding of conversations.
Towards this end, we develop a novel {\it \simmc ontology} that captures the detailed
multimodal interactions within dialog flows.
\sk{
Note that these dialog annotations are collected after the dialog 
data collection stage, with the help of professional linguists.
}
In this section, we describe the proposed \simmc ontology and the hierarchical labeling language centered around \textit{objects} 
(\refsec{subsec:annotation:ontology} and \ref{subsec:annotation:language}), and the multimodal coreference schema that links the annotated language with the co-observed multimodal context 
(\refsec{subsec:annotation:coref}).

\subsection{\simmc Annotation Ontology}
\label{subsec:annotation:ontology}
The \simmc ontology provides common semantics for both the assistant and user utterances. The ontology is developed in the Resource Development Framework (RDF) and is an expansion of the 
Basic Formal Ontology \cite{ontology_arp2015}.
It consists of four primary components:
\begin{itemize}[leftmargin=*,noitemsep,parsep=0pt,partopsep=0pt]
    \item \textbf{Objects:} A hierarchy of objects is defined in the ontology. This hierarchy is a rooted tree, with finer-grained objects at deeper levels.
    Sub-types are related to super-types via the \textit{isA} relationship, \eg,
    \textsc{sofa} \textit{isA} \textsc{furniture}.
    Fine-grained objects include \textsc{user}, \textsc{dress}, and \textsc{sofa}. 
    \item \textbf{\Activities:} A hierarchy of \activities are defined as a sub-graph of objects within the ontology. These represent \activities the virtual assistant can take like \textsc{get}, \textsc{refine}, and \textsc{add\_to\_cart}.
    \item \textbf{Attributes:} A given object has a list of attributes which relate that object to other objects, to primitive data types, or to enums. Finer-grained objects inherit the attributes of their parents. There are restrictions on the available types for both the domain and range of attributes. For example, a \textsc{sofa} can be related to a \textsc{company} via the \textit{brand} attribute. A \textsc{person} can be related to an item of \textsc{clothing} via the \textit{attentionOn} attribute. 
    The \textit{takesArgument} attribute relates \Activities and the objects they act upon.
    \item \textbf{Dialog Acts:} A hierarchy of dialog acts is also defined as a sub-graph of objects within the ontology. Dialog acts indicate the linguistically motivated purpose of the user or system’s utterance. They define the manner in which the system conveys information to the user and vice versa. Examples of dialog acts include: \textsc{ask}, \textsc{inform}, and \textsc{prompt}. Dialog acts are related to the \activities that they act upon via the \textit{takesArgument} attribute.
    \reftab{tab:simmc_flex_ontology} lists the \activities and dialog acts used in our work.
\end{itemize}

\subsection{\simmc Labeling Language}
\label{subsec:annotation:language}
From the \simmc ontology, we derive a compositional, linearized, and interpretable labeling language for linguistic annotation, allowing for the representation of the natural language utterances as well-formed subgraphs of the ontology \cite{kollar-etal-2018-alexa}. The labeling language consists of intents and slots \cite{Gupta+06}.
Intents are taken to represents instances of the types they are composed of and take one of two forms: 1) \textsc{dialog\_act:\activity:object} or 2) \textsc{dialog\_act:\activity:object}.attribute. Only combinations of objects and attributes declared to be valid in the ontology are made available in the labeling language. Within these intents, slots further specify values for attributes of objects, \activities, and attribute types.
In the basic case, slots take the form of attributes of the intent-level objects and restrict those attributes. More complex cases include slot-in-slot nesting to restrict the type of the embedding slot, object-attribute combinations for type-shifting contexts, \ie, utterances in which an intent-level object is identical to the range of another object's property, and a system of indexing to restrict objects introduced within the intent.
Crucially, the labeling language is speaker agnostic. It makes no distinction in the parses of the user's utterance versus those of the assistant.

A number of additional conventions are placed on the annotation task to ensure consistency and accuracy, which are detailed in Appendix \ref{sec:appendix_labeling}.
See \reftab{tab:simmc_furniture_ex_1} and \reftab{tab:simmc_fashion_ex_1} in Appendix \ref{sec:appendix_examples} for annotated dialog examples that show our \simmc ontology in action for both our datasets.

\subsection{\simmc Coreference Annotations}
\label{subsec:annotation:coref}
Note that the proposed labeling language allows for the annotation of object types in a dialog, which may in turn refer to specific canonical listings from the underlying multimodal contexts.
For example, given an annotated utterance ``[\textsc{da:request:get:chair} \textit{Show me the back of \underline{it}}]", the annotated object `\textsc{chair}' 
(\textit{\underline{it}}) would refer to a specific catalog item, represented as a item id within the image metadata.
To allow for structural grounding between the verbal and visual modalities in a shared catalog, we further annotate the mapping of object type mentions in the annotated utterance to the corresponding item id in the image metadata.
The final SIMMC annotations thus capture the semantic relations of objects in multimodal contexts with their corresponding dialog annotations (\activities, attributes and dialog acts), as outlined in the proposed SIMMC ontology (\refsec{subsec:annotation:ontology}).
We provide the detailed analysis of the datasets and the annotations in Appendix \ref{sec:dataset-analysis}.
\section{\simmc Tasks \& Metrics}
\label{sec:tasks_and_metrics}
We define several offline evaluation tasks within the \simmc framework to train conversational models on these new datasets using the fine-grained annotations that are provided.
We first provide the general offline evaluation framework for defining \simmc tasks (\refsec{subsec:tasks_and_metrics:formulation}),
and then present three major tasks that we focus on in this paper.
These are primarily aimed at replicating human-assistant actions in order to enable rich and interactive shopping scenarios (\refsec{subsec:tasks_and_metrics:tasks}).

\subsection{Offline Evaluation Framework}
\label{subsec:tasks_and_metrics:formulation}
Consider a generic \simmc dialog $\mathcal{D} = \{(U_i, A_i, M_i, a_i)\}_{i=1}^{N_r}$ 
that is $N_r$ rounds long, where $U_i$ and $A_i$ are the user and assistant
utterances, $M_i$ is the domain-specific multimodal context, 
and $a_i$ is the action (API call) taken by the assistant at round $i$, respectively.
Formally, a task is defined as:
At each round $t$, given the current user utterance $U_t$, the dialog history 
$H_t={(U_i, A_i)}_{i=1}^{t-1}$, multimodal context $M_t$, predict
the assistant action $a_t$ along with the free-form, natural language assistant response $A_t$.

\vspace{0.05in}
\begin{minipage}{0.60\textwidth}
    \centering
    \includegraphics[width=1.0\textwidth]{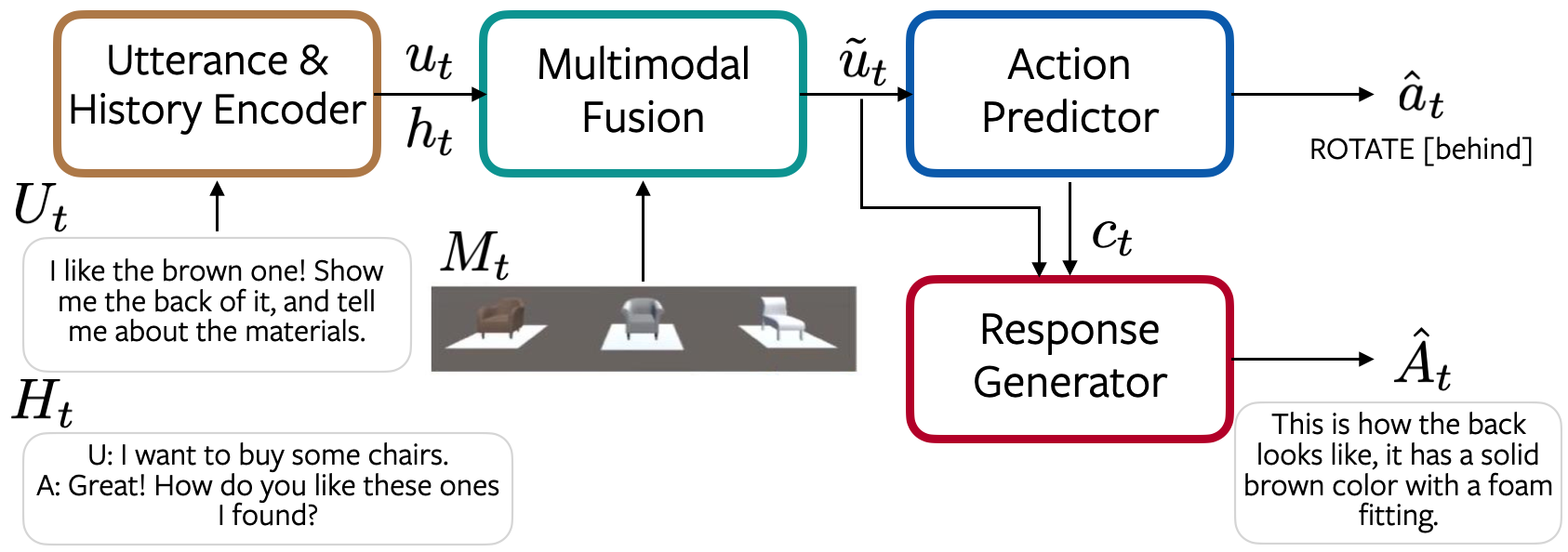}
    \captionof{figure}{
        \textbf{Assistant Model Architecture Overview} (\refsec{sec:models}):
        utterance and history encoder, multimodal fusion, action predictor, and response decoder. Example taken from \reffig{fig:teaser}. 
    }
    \vspace*{\captionReduceBot}
    \label{fig:model_overview_architecture}
\end{minipage}
\hspace{5pt}
\begin{minipage}{0.30\textwidth}
    \begin{center}
        \scalebox{0.75}{
        \setlength\tabcolsep{2pt}
        \begin{tabular}{p{0.25\columnwidth} p{0.95\columnwidth}}
        \toprule[\heavyrulewidth]
        {\hfil\textbf{Model}} & {\hfil\textbf{Functions}}\\
        \midrule
        HAE
        &$\displaystyle
            \begin{aligned}
                u_t = \text{LSTM}(U_t); h_t = \emptyset
            \end{aligned}
        $ \\
        \midrule
        HRE
        &$ \displaystyle
            \begin{aligned}
                u_t &= \text{LSTM}(U_t) \\
                \tilde{h}^{(i)}_t &= \text{LSTM}([U_i, A_i]) \\
                h_t &= \text{Attention}(u_t, [h^{(i)}_t]_{i=1}^{t-1})
            \end{aligned}
        $ \\
        \midrule
        MN
        &$ \displaystyle
            \begin{aligned}
                u_t &= \text{LSTM}(U_t) \\
                h_t &= \text{LSTM}([u_i]_{i=1}^{t-1})
            \end{aligned}
        $ \\
        \midrule
        T-HAE
        &
        $\displaystyle
            \begin{aligned}
                u_t = \text{Transformer}(U_t); h_t = \emptyset
            \end{aligned}
        $
        \\
        \bottomrule[\heavyrulewidth]
        \end{tabular}
        }
    \end{center}
    \vspace*{\captionReduceTop}
    \captionof{table}{Overview of \simmc models. 
    See \refsec{sec:models} for details.}
    \label{tab:model_components}
    \vspace*{\captionReduceBot}
\end{minipage}

\vspace{0.05in}
\subsection{\simmc Tasks}
\label{subsec:tasks_and_metrics:tasks}
The proposed offline evaluation framework has a three-fold advantage:
(a) It accurately represents the scenario encountered by a \simmc model during 
deployment. 
In other words, models trained for the above task can be deployed 
to interact with humans to provide a situated, interactive, multimodal conversation.
(b) Instead of evaluating the performance on the entire dialog, we evaluate models on a per-turn basis with the ground-truth history.
This avoids taking the conversation out of the dataset and reduces the dependency
on a user simulator, with the caveat of not encouraging the model to be able to learn multiple equally valid routes to satisfy the user's request.
(c) Finally, it facilitates us to define and evaluate several sub-tasks such as action
prediction, response generation, and dialog state tracking, within \simmc, which allows us to bootstrap from prior work on these sub-tasks.


\noindent \textbf{Task 1: Structural API Call Prediction.} This task involves predicting the assistant action 
$a_t$ as an API call along with the necessary arguments, 
using $H_t, M_t, U_t$ as inputs.
For example, enquiring about an attribute value (\eg, price) for a shared
furniture item is realized through a call to the \textit{SpecifyInfo} API with
the \textit{price} argument.
A comprehensive set of APIs for our \simmc dataset is given in 
\reftab{tab:api_call_list}.
Apart from these APIs, we also include a \textit{None} API call to catch situations without 
an underlying API call, \eg., a respond to \textit{`U: Can I see some tables?'} as
\textit{`A: What color are you looking for?'} does not require any API calls.
Action prediction is cast as a round-wise, multiclass classification problem over the
set of APIs, measured using $1-0$ accuracy of predicting the action 
taken by the assistant during data collection.
However, we note that there could be several actions that are equally 
valid in a given context.
For instance, in response to \textit{`U: Show me some black couches.'}, one could 
show black couches \textit{`A: Here are a few.'} or enquire further about specific preferences
\textit{`A: What price range would you like to look at?'}.
Since accuracy does not account for the existence of multiple valid actions, 
we use perplexity (defined as the exponential of the mean log-likelihood) alongside accuracy. 
To also measure the correctness of the predicted action (API) arguments, we use attribute accuracy
compared to the collected datasets.

\noindent
\textbf{Task 2: Response Generation.} This task measures the relevance of the assistant response $A_t$ in the current turn. We evaluation in two ways, as a:
(a) Conditional language modeling problem, where the closeness between the generated and
ground-truth response is measured through using BLEU-4 score \cite{papineni2002bleu}, and,
(b) Retrieval problem, where performance of the model to retrieve the ground-truth response from
a pool of $100$ candidates (randomly chosen unique to each turn) is measured using standard
retrieval metrics like recall@k ($k=1,5,10$), mean rank, and mean reciprocal rank.

\noindent
\textbf{Task 3: Dialog State Tracking (DST)}.
The dialog annotations collected using the flexible ontology
enable us to study dialog state tracking (DST) in \simmc, aside from providing additional supervision to train goal-driven agents.
As mentioned in \refsec{sec:annotations}, the user and assistant utterances are accompanied with 
a hierarchy of 
\textit{dialog act} labels and text spans for the corresponding slots or attributes, if any.
The goal of DST is to systematically track the dialog acts and the associated slot pairs across multiple turns. We use the intent and slot prediction metrics (F1), following prior work in DST \cite{dstc2}.

\section{Modeling for \simmc Tasks}
\label{sec:models}
We now propose several models building on top of prior work and train them on the tasks 
formulated in \refsec{sec:tasks_and_metrics} to benchmark the \simmc dataset.
We define two classes of models for the \simmc tasks: (1) Assistant models, which aims at mimicking the assistant actions and responses (Task 1 \& 2), and (2) User belief tracking model (Task 3) that output semantic parses of user utterances, agnostic of future assistant actions.
Our principal Assistant model architecture is illustrated in 
\reffig{fig:model_overview_architecture}, which is composed of  four main components:
Utterance and History Encoder, MultiModal Fusion, Action Predictor, and Response Generator. 
On the other hand, our user belief model builds upon the state-of-the-art DST models and extend them to accommodate for multimodal input. 
Inspired by \cite{simpletod}, we adapt one such model and finetune a pretrained GPT-2 language 
model \cite{gpt2} to both action prediction and belief tracking.


\paragraph{Utterance \& History Encoder.}
\label{subsec:models:encoder}
The utterance and history encoder takes as input the user utterance at the current round $U_t$ and the dialog history so
far $H_t$, to produce the utterance encoding $u_t$ and history encoding $h_t$ to
capture the respective textual semantics.
Inspired from prior work, we consider several utterance and history encoders, whose
functional forms are outlined in \reftab{tab:model_components}.
We embed each token in the input sequences ($U_t$ or $H_t$) through learned word
embeddings of size $D_W$, which are further fed into the encoders.
These output $u_t \in \mathbb{R}^{N_U \times D_H}$ and 
$h_t \in \mathbb{R}^{t-1 \times D_H}$, where
$D_H$ is the embedding size, $N_U$ is length of the user utterance.
\textbf{(a) History-Agnostic Encoder (HAE)} ignores the dialog context $H_t$ to 
only encode the user utterance through an LSTM \cite{lstm} for the downstream
components.
\textbf{(b) Hierarchical Recurrent Encoder (HRE)}\cite{hre_serban} models dialogs
at two hierarchical recurrence levels of utterance and turn.
The utterance encoder LSTM operates at the former, while a history LSTM that
consumes the hidden states of utterance encoder LSTM from all the previous rounds
($[u_i]_{i=1}^{t-1}$) operates at the latter.
\textbf{(c) Memory Network (MN)} encoder \cite{memory_sukhbaatar15} treats
dialog history $H_t$ as a collection of memory units comprising user and assistant
utterance pairs concatenated together, and
uses the current utterance encoding $u_t$ to selectively attend to these
units to produce the utterance-conditioned history encoding $h_t$.
\textbf{(d) Transformer-based History-Agnostic Encoder (T-HAE)} is a variant of 
HAE with the LSTMs replaced with Transformer units \cite{attention_vaswani2017}
that achieved state-of-the-art results in language modeling \cite{bert_devlin19}.

\paragraph{Multimodal Fusion.}
This component fuses semantic information from the 
text ($u_t$ and $h_t$) and the multimodal context $M_t$ (described in \refsec{sec:experiments}), to create the fused
context tensor $\tilde{u}_t \in \mathbb{R}^{N_U \times 2D_H}$, which is 
double the size of $u_t$ in the last dimension.
In our setup, the multimodal context is modelled as a tensor of size 
$M_t \in \mathbb{R}^{N_M \times D_M}$, where $N_M$ is the number of multimodal units 
for the current round $t$ and $D_M$ is the multimodal embedding size.
Note that all of our models have the same architecture to fuse multimodal information.
At a high level, we first embed $M_t$ to match its size to $D_H$ using a linear layer
followed by a non-linearity (ReLU) (\refeq{eq:mf_embed}),
then use the utterance encoding $u_t$ to attend to the multimodal units 
(\refeq{eq:mf_attend}),
and finally fuse the attended multimodal information with $\tilde{u}_t = [u_t ; m_t]$.
More concretely,
\noindent
\begin{minipage}{.49\linewidth}
\begin{equation}
  \tilde{M}_t = \text{Tanh}(\text{Linear}(M_t)), \label{eq:mf_embed}
\end{equation}
\end{minipage}
\begin{minipage}{.49\linewidth}
\begin{equation}
  m_t = \text{Attention}(u_t, \tilde{M}_t, \tilde{M}_t), \label{eq:mf_attend}
\end{equation}
\end{minipage}

\noindent where Attention operator for a query $Q$ over the key $K$
(of size $D_K$) and value $V$ is defined as
\begin{align}
    \text{Attention}(Q, K, V) = \text{Softmax}\left(\frac{QK^T}{\sqrt{D_K}}\right)V.
    \label{eq:attention_def}
\end{align}

\paragraph{Action Predictor.}
Using the fused context $\tilde{u}_t$, the Action Predictor predicts 
the appropriate action (API) $\hat{a}_t$ and the corresponding API arguments
to be taken by the assistant.
The former is a multi-class classification that chooses from a set of actions (APIs) 
while the latter is a multi-way classification modelled as a set of binary classifiers
one for each attribute like \textit{category, color, price}, \etc.
For a list of APIs and their arguments supported in our work, see \reftab{tab:api_call_list} in Appendix \ref{sec:appendix_api_call_list}.
First, the tensor $\tilde{u}_t$ is transformed into a vector 
$q_t \in \mathbb{R}^{N_H}$ through self-attention through the
attention parameter $\theta_{\text{AP}}$ (\refeq{eq:ap_selfattention}).
Next, we learn a classifier (MLP) that takes in $q_t$ to predict the distribution over 
the possible APIs (\refeq{eq:ap_classify}).
In addition, we also learn several binary classifiers (MLP) one each for the
corresponding API arguments.
Having predicted the structured API calls, we execute them and encode the
output as action context $c_t \in \mathbb{R}^{N_A \times D_A}$, where
$N_A$ is the number of context units and $D_A$ is action context embedding size.
The dataset-dependent specifics about the API call output encoding $c_t$ are in
\refsec{sec:dataset_specific_features}.
Finally, $c_t$ and $\tilde{u}_t$ feed into the last component to
generate the assistant response.
As the training objective, we minimize the cross entropy loss $\mathcal{L}_{a}$
for both the action and action attributes.

\noindent
\begin{minipage}{.49\linewidth}
\begin{equation}
  q_t = \text{Attention}(\theta_{\text{AP}}, \tilde{u}_t, \tilde{u}_t) .
    \label{eq:ap_selfattention}
\end{equation}
\end{minipage}
\begin{minipage}{.49\linewidth}
\begin{equation}
  p(\hat{a}_t | U_t, M_t) = \text{Softmax}(\text{Linear}(q_t)). \label{eq:ap_classify}
\end{equation}
\end{minipage}

\paragraph{Response Generator.} 
\label{subsec:models:response_generator}
As the last component, the response generator (decoder) generates the assistant response $\hat{A}_t$.
In our work, we model it as a language model conditioned on both $c_t$ and $\tilde{u}_t$.
The former ensures that the response is influenced by the API call output while the 
latter maintains semantic relevance to the user utterance.
For example, the response to \textit{`Show me black couches less than \$500'}
depends on the availability of such couches in the inventory and could lead to either 
\textit{`Here are some'} or \textit{`Sorry, we do not have any black couches cheaper than \$500'}.
For models that use LSTM for user and history encoders, the response decoder is
also an LSTM with attention over fused context $\tilde{u}_t$ and action API output
$c_t$ at every decoding time step, similar to \cite{bahdanau_2014}.
Similarly, we use a Transformer-based decoder for the other models to ensure consistent
underlying architecture (either LSTM or transformer).
Like any conditional language model, we decode individual tokens at each time step to
generate $\hat{A}_t$, and minimize the negative loglikelihood $\mathcal{L}_A$ 
of the human response under the model during training.

\paragraph{Dialog State Tracking (DST).} 
\label{subsec:models:dst}
In contrast to the assistant model that mimics the assistant actions and responses (Task 1 \& 2), the user belief model aims to output semantic parses of user-side dialog ($b_t$), strictly given the multimodal contexts available to the user, and agnostic of future assistant actions.
Thus, $b_t = \text{DST}(U_t, H_t, M_t)$.
We utilize the state-of-the-art DST models from the recent literature: TRADE \cite{trade}, which implements a pointer network that generates text spans for each slot, and an approach similar to SimpleTOD \cite{simpletod}, which fine-tunes the pre-trained GPT-2 language model to output the user belief state labels \cite{gpt2}.

In addition, we extend the SimpleTOD model to allow for multimodal input (SimpleTOD+MM).
Specifically, we cast the belief tracking problem as a causal language modeling task, where belief labels and multimodal contexts are represented as additional tokens.
A single training sequence can then be represented as the concatenation of input and target output $y_t = [H_t; M_t; U_t; b_t]$, where both $M_t$ and $b_t$ are cast as string tokens of key value pairs.
The language model is then fine-tuned to learn the joint probability with $p(x_t) = \prod_{i=1}^n p(x_{t,i}|x_{t,<i})$ for all $n$ tokens in a sequence.
At inference time, we provide the user input context $x_t = [H_t; M_t; U_t]$ as a seed for the language model, and parse the generated output to obtain the structural representation of user belief states.

We further extend the SimpleTOD+MM model to Tasks (1) and (2) by adding actions and assistant responses to the concatenation of input and target output, \ie, $[H_t; M_t; U_t; b_t; a_t; A_t]$. At test time, we provide the input context plus oracle belief state $[H_t; M_t; U_t; b_t]$ and parse the generated response to extract the action and system utterance. We refer to this model as STOD++.
\section{Experiments \& Results}
\label{sec:experiments}

\paragraph{Dataset Splits and Baselines.}
Our models are learned on randomly sampled \texttt{train} ($60\%$), model
hyperparameters chosen via early stopping using performance on 
\texttt{dev} ($10\%$),
and evaluation numbers reported on the unseen \texttt{testdev} ($15\%$).
In addition to the models described in \refsec{sec:models},
we consider two simple baselines that use TF-IDF features for utterance
and history encoders for action prediction, and LSTM-based language model (LSTM)
trained solely on assistant responses, and compare against them.

\paragraph{Dataset-specific Model Details.}
\label{sec:dataset_specific_features}
We provide details around modeling multimodal context $M_t$ and encoding action (API call) 
output $c_t$ for each of the \simmc datasets below.

\textbf{A. \furniturefull.} 
Since the data collection for \furniture is
grounded in a co-observed virtual 3D environment (\refsec{sec:dataset}),
its state becomes the multimodal context $M_t$.
For both \textit{carousel} and \textit{focused} environment states,
we concatenate the furniture item representation in the corresponding 
slot (or zero vector if empty) with its positional embedding
(\textit{`left', `center', `right', `focused'}) that are jointly learned, to give 
$M_t \in \mathbb{R}^{N_M\times D_M}$ with $N_M=3$ (carousel) or $N_M=1$ (focused).
In addition, each furniture item is represented with the concatenated GloVe 
embeddings~\cite{glove_pennington2014} of its attributes like category, color, intended room, \etc.
Similarly, we construct the action output $c_t \in \mathbb{R}^{N_A \times D_A}$ using the
environment representation after executing the necessary structural API call, \eg., \textit{search} for an item or \textit{focus} on an existing item.
The information seeking action \textit{SpecifyInfo} is an exception, for which $c_t$ is
the GloVe embedding of the attributes of the desired item.

\textbf{B. \fashionfull.}
Dialogs in \fashion use a fashion item (updated as the conversation progresses) and 
a sequence of `previously viewed items' (memory) as context (\refsec{subsec:dataset:simmc_furniture}).
To reflect this scenario, we extract the representations for each fashion item using
concatenated GloVe embeddings of its attributes (similar to \furniture) in addition to learning
the source embedding (\textit{`memory'} or \textit{`current'} item), as the multimodal 
context $M_t \in \mathbb{R}^{4\times D_M}$.
Akin to \furniture, $c_t$ is modeled simply as the updated multimodal state $M_t$ after executing
the current API.

\paragraph{Supervision.}
We learn \simmc models end-to-end by jointly minimizing the sum of
the action prediction and the response generation losses, \ie., $\mathcal{L}_a + \mathcal{L}_A$.
To extract supervision for API call prediction (along with attributes), we utilize both
the assistant (Wizard) interface activity during data collection
(\refsec{sec:item_scene_data}) and the fine-grained NLU annotations.
Our implementation details are in Appendix \ref{sec:appendix_implementation_details}.

\paragraph{Results.}
\reftab{tab:results_brief} summarizes the performance of \simmc Assistant models on
structural API prediction and response generation.

The key observations are:
(a) All \simmc neural models (HAE, HRE, MN, T-HAE) outperform the baselines (TF-IDF and LSTM) across all metrics for both the datasets.
(b) HRE consistently achieves the highest API prediction accuracy 
for \furniture ($80.0\%$, jointly with HAE) and \fashion 
($81.9\%$, jointly with HAE and MN).
STOD++ achieves $61.4\%$ accuracy on attributes, an overwhelming ~$7\%$ point improvement over T-HAE for \furniture, benefiting from
having access to the oracle belief state where the user requested attributes
are formally represented.
(c) For response generation, HRE has superior BLEU score for \furniture and HRE for \fashion.
Surprisingly, T-HAE has the least BLEU scores amongst \simmc models perhaps
due to resorting to safe, frequent

\begin{minipage}{0.63\textwidth}
      \scalebox{0.7}{
        \begin{tabular}{
            p{0.13\columnwidth}
            ccccccccc
        }
        \toprule[\heavyrulewidth]
        \multirow{2}{*}{\textbf{Model}}
        & \multicolumn{3}{c}{\textbf{Task 1. API Prediction}}
        & \multicolumn{6}{c}{\textbf{Task 2. Response Generation}} \\
        \cmidrule(r){2-4}
        \cmidrule(r){5-10}
            & Acc$\uparrow$ 
            & Perp$\downarrow$
            & A.Acc$\uparrow$
            & BLEU$\uparrow$ 
            & r@1$\uparrow$ 
            & r@5$\uparrow$
            & r@10$\uparrow$
            & Mean$\downarrow$
            & MRR$\uparrow$\\
        \midrule
        \multicolumn{10}{c}{\furniture}\\
        \midrule
        TF-IDF
            & \reportvalbrief{77.1}{0}
            & \reportvalbrief{2.59}{0}
            & \reportvalbrief{57.5}{0}
            & -
            & -
            & - 
            & - 
            & - 
            & -\\
        
        LSTM 
            & -
            & -
            & -
            & \reportvalbrief{0.022}{0}
            & \reportvalbrief{4.1}{0}
            & \reportvalbrief{11.1}{0}
            & \reportvalbrief{17.3}{0}
            & \reportvalbrief{46.4}{0}
            & \reportvalbrief{0.094}{0}\\
        
        HAE
            & \textbf{\reportvalbrief{79.7}{0}}
            & \reportvalbrief{1.70}{0}
            & \reportvalbrief{53.6}{0}
            & \reportvalbrief{0.075}{0}
            & \reportvalbrief{12.9}{0}
            & \reportvalbrief{28.9}{0}
            & \reportvalbrief{38.4}{0}
            & \reportvalbrief{31.0}{0}
            & \reportvalbrief{0.218}{0}\\

        HRE 
            & \textbf{\reportvalbrief{80.0}{0}}
            & \textbf{\reportvalbrief{1.66}{0}}
            & \reportvalbrief{54.7}{0}
            & \reportvalbrief{0.075}{0}
            & \reportvalbrief{13.8}{0}
            & \reportvalbrief{30.5}{0}
            & \reportvalbrief{40.2}{0}
            & \reportvalbrief{30.0}{0}
            & \reportvalbrief{0.229}{0}\\
        
        MN 
            & \reportvalbrief{79.2}{0}
            & \reportvalbrief{1.71}{0}
            & \reportvalbrief{53.3}{0}
            & \textbf{\reportvalbrief{0.084}{0}}
            & \textbf{\reportvalbrief{15.3}{0}}
            & \textbf{\reportvalbrief{31.8}{0}}
            & \textbf{\reportvalbrief{42.2}{0}}
            & \textbf{\reportvalbrief{29.1}{0}}
            & \textbf{\reportvalbrief{0.244}{0}}\\
        
        T-HAE 
            & \reportvalbrief{78.4}{0}
            & \reportvalbrief{1.83}{0}
            & \reportvalbrief{53.6}{0}
            & \reportvalbrief{0.044}{0}
            & \reportvalbrief{8.5}{0}
            & \reportvalbrief{20.3}{0}
            & \reportvalbrief{28.9}{0}
            & \reportvalbrief{37.9}{0}
            & \reportvalbrief{0.156}{0}\\
 
         STOD++$^\dagger$ 
            & \reportvalbrief{72.2}{0}
            & -
            & \textbf{\reportvalbrief{61.4}{0}}
            & \reportvalbrief{0.155}{0} 
            & -
            & - 
            & - 
            & -
            & -\\

        \midrule
        \multicolumn{10}{c}{\fashion}\\
        \midrule
        TD-IDF
            & \reportvalbrief{78.1}{0}
            & \reportvalbrief{3.51}{0}
            & \reportvalbrief{57.9}{0}
            & -
            & -
            & -
            & -
            & -
            & -\\
        
        LSTM
            & -
            & -
            & -
            & \reportvalbrief{0.022}{0}
            & \reportvalbrief{5.3}{0}
            & \reportvalbrief{11.4}{0}
            & \reportvalbrief{16.5}{0}
            & \reportvalbrief{46.9}{0}
            & \reportvalbrief{0.102}{0}\\
        
        HAE
            & \textbf{\reportvalbrief{81.0}{0}}
            & \textbf{\reportvalbrief{1.75}{0}}
            & \reportvalbrief{60.2}{0}
            & \reportvalbrief{0.059}{0}
            & \reportvalbrief{10.5}{0}
            & \reportvalbrief{25.3}{0}
            & \reportvalbrief{34.1}{0}
            & \reportvalbrief{33.5}{0}
            & \reportvalbrief{0.190}{0}\\
        
        HRE 
            & \textbf{\reportvalbrief{81.9}{0}}
            & \textbf{\reportvalbrief{1.76}{0}}
            & \textbf{\reportvalbrief{62.1}{0}}
            & \textbf{\reportvalbrief{0.079}{0}}
            & \textbf{\reportvalbrief{16.3}{0}}
            & \textbf{\reportvalbrief{33.1}{0}}
            & \textbf{\reportvalbrief{41.7}{0}}
            & \textbf{\reportvalbrief{27.4}{0}}
            & \textbf{\reportvalbrief{0.253}{0}}\\
        
        MN 
            & \textbf{\reportvalbrief{81.6}{0}}
            & \reportvalbrief{1.74}{0}
            & \reportvalbrief{61.6}{0}
            & \reportvalbrief{0.065}{0}
            & \textbf{\reportvalbrief{16.1}{0}}
            & \reportvalbrief{31.0}{0}
            & \reportvalbrief{39.4}{0}
            & \reportvalbrief{29.3}{0}
            & \reportvalbrief{0.245}{0}\\

        T-HAE 
            & \reportvalbrief{81.4}{0}
            & \textbf{\reportvalbrief{1.78}{0}}
            & \textbf{\reportvalbrief{62.1}{0}}
            & \reportvalbrief{0.051}{0}
            & \reportvalbrief{10.3}{0}
            & \reportvalbrief{23.2}{0}
            & \reportvalbrief{31.1}{0}
            & \reportvalbrief{37.1}{0}
            & \reportvalbrief{0.178}{0}\\
        \bottomrule[\heavyrulewidth]
        \end{tabular}
    }
    \vspace*{-10pt}
    \captionof{table}{
        Results for:
        \textbf{(1) API prediction} via \underline{acc}uracy, \underline{perp}lexity and \underline{a}ttribute \underline{acc}uracy, and,
        \textbf{(2) Response generation} via \underline{BLEU}, 
        \underline{r}ecall\underline{@k} (k=1,5,10), \underline{mean} rank, and mean
        reciprocal rank (\underline{MRR}).
        Std Errors: $<0.5\%$ for Acc, A.Acc, r@1, r@5, r@10, mean;
        $0.005$ for BLEU and MRR.
        $^\dagger$Uses oracle belief state.
        }
    \label{tab:results_brief}
\end{minipage}
\hspace{8pt}
\begin{minipage}{0.31\textwidth}
    \begin{center}
        \scalebox{0.8}{
        \setlength\tabcolsep{3.5pt}
        \begin{tabular}{lcc}
        \toprule[\heavyrulewidth]
        \multicolumn{1}{c}{\multirow{2}{*}{\textbf{Model}}} & \multicolumn{2}{c}{\textbf{T3. DST}} \\
        \cmidrule(r){2-3}
         & In.F1$\uparrow$  & Sl.F1$\uparrow$  \\
        \midrule
        \multicolumn{3}{c}{\furniture}  \\
        \midrule
        TRADE & - & 45.5 \\
        SimpleTOD & \textbf{75.0} & 50.1 \\
        SimpleTOD+MM & 74.1 & \textbf{60.2} \\
        \midrule
        \multicolumn{3}{c}{\fashion}  \\
        \midrule
        TRADE & - & 32.8  \\
        SimpleTOD & 56.5 & 37.3 \\
        SimpleTOD+MM & \textbf{59.1} & \textbf{43.5} \\
        \bottomrule[\heavyrulewidth]
        \end{tabular}
        }
    \end{center}
    \captionof{table}{
        Results for: \textbf{(3) Dialog State Tracking (DST)}, measured with \underline{In}tent and \underline{Sl}ot prediction F1 metrics.
        $\uparrow$: higher is better, $\downarrow$: lower is better.
        Bold denotes the best for each metric.
    }
    \label{tab:results_dst}
\vspace{-10pt}
\end{minipage}




\vspace{12pt}

\noindent responses.
(d) The confusion matrix for HRE on \furniture 
(Appendix \ref{sec:appendix_model_visualization}) reveals a high confusion
between \textit{SearchFurniture} and \textit{None}.

This is intuitive as searching for an item or further obtaining user preferences to narrow the
search are equally valid actions for their context.
Note that the proposed assistant models do not leverage the rich, fine-grained annotations
of the \simmc datasets (understandably so) as they are adaptations
of existing state-of-the-art models.

\reftab{tab:results_dst} presents the performance of the state-of-the-art DST models on the \simmc datasets.
It can be seen that the pretrained GPT-2 based SimpleTOD models outperform the TRADE baseline.
Note that the original TRADE implementation does not include the dialog act prediction, hence it is not reported here as well. 
When the multimodal contexts are added as input (SimpleTOD+MM), the performance improves upon the text-only SOTA model (SimpleTOD) on both datasets, especially in the slot prediction metrics. This demonstrates the efficacy of grounding the multimodal contexts for DST, by better resolving the multimodal coreferences. 
In general, the performance on the \fashion dataset is typically better than on the \furniture dataset.
This could be due to the nature of the dialogs in the \fashion dataset, which involves more natural and casual utterances, as evident in the lower annotator agreement as well (Appendix \ref{sec:appendix_annotation_process}).

\section{Conclusion}
In this work, we presented {\bf Situated Interactive Multi-Modal Conversations (SIMMC)}, an important new direction towards building next generation virtual assistants with evolving multimodal inputs.
In particular, we collected two new datasets using the \simmc platform, and provided the contextual NLU and coreference annotations on these datasets, creating a new SIMMC task for the community to study.
We established several strong baselines for some of the tasks enabled by the datasets, showcasing various uses of the datasets in real-world applications.
The fine-grained annotations we collected open the door for studying several different tasks in addition to the ones highlighted in this work, which we leave as future work for the community to tackle. 

\section*{Acknowledgements}
\vspace{-.1in}
We thank Pararth Shah, Oksana Buniak, Semir Shafi, {\"U}mit Atlamaz, Jefferson Barlew, Becka Silvert, Kent Jiang, Himanshu Awasthi, and Nicholas Flores for their invaluable technical contributions to the data collection platforms, annotation schema development, annotation process, tooling and coordination. We also extend many thanks to all the annotators who meticulously labelled these datasets.

\bibliography{bibliography}
\bibliographystyle{coling}
\vfil
\pagebreak

\normalsize
\begin{appendix}
    \appendix
\appendix


\section{Dataset \& Annotation Analysis}
\label{sec:dataset-analysis}

\begin{figure*}[t]
    \centering
    \includegraphics[width=\textwidth]{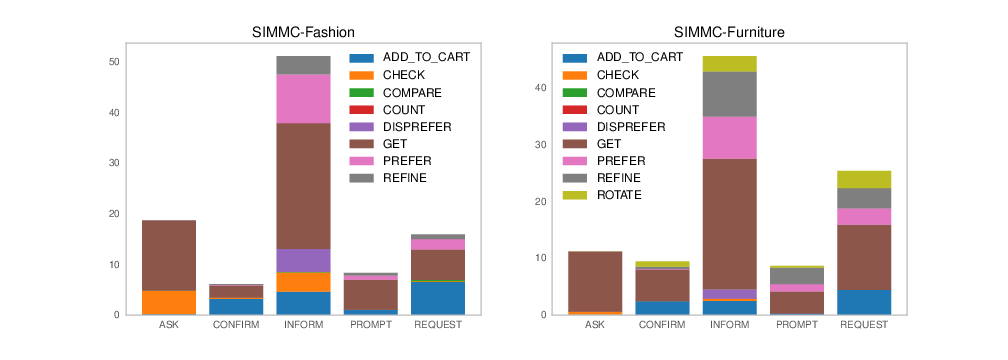}
    \vspace*{\captionReduceTop}
    \caption{
        Distribution of Dialog Acts and \Activities in the \simmc datasets. 
        See \refsec{sec:dataset-analysis} for details.
    }
    \vspace*{\captionReduceBot}
    \label{fig:nlu_annotation_distr}
\end{figure*}

\paragraph{Annotation Analysis.}
Using the unified ontology framework described in 
\refsec{subsec:annotation:ontology}, we annotate both the user and assistant
utterances of the \simmc datasets.
There are effectively $5$ dialog acts which are respectively combined with $9$ \activities for (\furniture) and $8$ \activities for (\fashion); the latter by design excludes \textsc{count} and
\textsc{rotate}.
A detailed list with examples is in Appendix, \reftab{tab:simmc_flex_ontology}.
Not all combinations of dialog acts and \activities are observed in our dataset, \ie,
about $38/45$ for \furniture and $32/40$ for \fashion respectively.
For instance, a \textsc{request:disprefer} utterance is an invalid combination.
The key takeaways from \reffig{fig:nlu_annotation_distr} are:
(a) \textsc{inform} is the most
dominant dialog act ($50\%$ in \fashion and $45\%$ in \furniture).
This is intuitive as conversations in shopping domain require the
user to \textit{inform} the assistant of their preferences, 
while the assistant \textit{informs} the user about the item attributes and 
availability.
(b) Interestingly, \textsc{get} is the dominant \activity across most dialog acts,
where the assistant either \textit{gets} new items or additional information about
existing items that the user is perusing.
(c) The relatively low occurrence of the \textsc{confirm} dialog act perhaps arises 
from the effectiveness of the human assistant agent. 
This is desirable to avoid learning assistant models that excessively repeat user
requests, \eg, repeatedly seek explicit confirm, as this leads to lower user satisfaction.
Note that this analysis of the dialog act and \activity distribution is per 
sentence, with an utterance occasionally containing multiple sentences 
(see \reffig{fig:teaser} for an example).

\paragraph{User Satisfaction Metrics.}
Since \simmc datasets aim at goal-oriented dialog, we also collect turn-level and
dialog-level user satisfaction scores in the range of 1-5 as part of the data collection.
The dialog-level user satisfaction scores for the \furniture dataset average at
$4.69 \pm 0.77$, showing a heavy concentration around 5.
Since the dialogs are collected between humans interacting with each other, we hypothesize that the
the assistant (wizard) is able to efficiently respond to user requests, leading to high
satisfaction scores.
Similar trends were observed across different metrics for both datasets.
Therefore, we drop further analysis on this front due to the absence of a clear 
signal in these collected metrics.



\section{Details for \simmc Labeling Language}
\label{sec:appendix_labeling}


A number of additional conventions are placed on the annotation task to ensure consistency and accuracy.

\noindent
\textbf{Type ambiguity.} When an object appears in an utterance, the most fine-grained type is annotated. For example, in the utterance ``\textit{Show me some dresses}", the token `dresses' needs to be annotated as \textsc{dress}, as opposed to a coarser-grained type \textsc{clothing}. When more than one fine-grained type is possible, the annotator utilizes a parent-level coarse-grained type instead. Thus the assigned type is the finest-grained type that still captures the ambiguity. 

\noindent
\textbf{Attribute ambiguity.} Attributes are annotated when they are unambiguous. When there is uncertainty in the attribute that should be selected for the representation, the annotator falls back to a more generic attribute. For example when asking about an \textsc{furniture} item the user may specify a particular dimension, e.g. \textit{I want a couch that is 2 feet wide}. In this case, \textit{2 feet} can be annotated with the specific attribute \textit{width}. However, if the dimension is not specified, the more general attribute \textit{dimensions} would be used.

\noindent
\textbf{Attribute inverses.} When an attribute can be annotated in two different directions, a canonical attribute is defined in the ontology and used for all annotations. For example, \textit{attentionOn} and \textit{inAttentionOf} are inverses. The \textsc{user} object is connected to \textsc{furniture} or \textsc{clothing} objects via \textit{attentionOn} if the user is looking at an instance of these objects. Inverserly, those same objects are connected to the \textsc{user} object via the \textit{inAttentionOf} attribute. The former is designated as the canonical attribute in this case, and used for labeling purposes.

\noindent
\textbf{Smart prefixes.} Attribute slots are prefixed by \textit{A} and \textit{O} respectively to indicate whether they serve to restrict the intent-level \Activity or Object. This is primarily for human-annotator convenience. For example the attribute \textit{amount} is an attribute of the Activity \textsc{GET}. The attribute \textit{color} is an attribute of \textsc{clothing}. Annotating an assistant response like \textit{'I found five green dress'} yields spanning \textit{five} with \textsc{a}.amount and \textit{green} with \textsc{o}.color.

\noindent
\textbf{Attribute variables.} The attribute \textit{.info} is employed when the speaker's
intent targets more than one attribute simultaneously.
The specific attributes being targeted are then identified with the INFO smart prefix. For example, \textit{'What is the color and brand of this skirt?'} is annotated with the intent \textsc{da:ask:get:skirt}.info and the tokens \textit{color} and \textit{brand} are labeled as \textsc{info}.color and \textsc{info}.brand respectively.

\section{Details for NLU/NLG/Coref Data Annotation}
\label{sec:appendix_annotation_process}

Data were annotated in two stages: (1) NLU/NLG followed by (2) image-based coreference annotations. 

During the NLU/NLG stage, annotators were provided full dialog context for a single dialog and asked to annotate both the user and assistant's utterances. Image context was not available, and annotators were instructed to use dialog context only up to the target utterance. Essentially all (98.4\%) annotations were single-annotated by annotators who passed an evaluation test; while 1.6\% were double-annotated.  In cases of disagreement between two annotators, a third annotator either selected one of the proposed annotations, or overrode both with a new one.

In order to estimate and improve quality, we double-annotated an additional 12.6\% of the data after the fact and applied two measures of inter-annotator agreement: exact matches between semantic parses, and a modified F1 score. 50\% of fashion and 60\% of furniture annotations were exact matches. Given sample sizes, this corresponds respectively to 95\% confidence intervals of 49-50\% and 58-63\%. In contrast to this binary exact measure, the F1 measures can assume values between 0 and 1. Using this measure, furniture annotations were 73.8\% similar while fashion annotations were 72.5\% similar .

During image-based coreference, annotators were provided all dialog and image context up until the turn in question. Review of the process suggested it was easy enough for the high-skilled pool annotators to perform without quality checks. By their own account, annotators self-reported 98\% confidence in their decision to link an object to an intent; and 98\% confidence in their specific choice of object given a link was required. 

\reftab{tab:object_attributes}  presents the Object Classes that were made available to the annotators for annotation as well as the attributes of these Classes. Attributes are listed alphabetically and type information is provided. Note that for readability attributes derived via inheritance from supertype to subtype are not repeated.
Classes that were exposed to annotators but had no attributes are not presented here. Attribute ambiguity is indicated by indenting.

\reftab{tab:activity_attributes} presents the Activity Classes that were made available to the annotators for annotation as well as the attributes of these Classes (see \reftab{tab:simmc_flex_ontology} for examples and definitions). Type information and attributes are provided. Note that for readability attributes derived via inheritance from supertype to subtype are not repeated. All Activities had the attributes amount an INTEGER, endTime and startTime (\texttt{DATE\_TIME}s). Only Activities with additional attributes are listed below.

\begin{table*}[t]
    \begin{center}
        \setlength\tabcolsep{5pt}
        \begin{tabular}{p{0.15\textwidth}p{0.75\textwidth}}
        \toprule[\heavyrulewidth]
        \underline{CLOTHING}&
            ageRange, amountInStock, availableSizes, brand, clothingCategory, clothingStyle, color, condition, customerRating, embellishment, forGender, forOccasion, forSeason, itemDescription, madeIn, material, ordinal, pattern, price, sequential, size, items, soldBy, warmthRating, waterResistance \\
        \underline{COMPANY}&
            headquarteredIn, name*, ordinal, sequential \\
        \underline{DATE\_TIME}&
            date, month, time, week, weekday, year \\
        \underline{DISPLAY}&
            displayPostion (displayFirst, displaySecond, displayThird) \\
        \underline{DRESS}&
            dressStyle, hemLength, hemStyle, necklineStyle, sleeveLength, sleeveStyle, waistStyle \\
        \underline{EVENT}&
            duration, elapsedTime, endTime, eventType, hasPart, name, remainingTime, startTime \\
        \underline{FURNITURE}&
             ageRange, amountInStock, assemblyRequired, brand, color, condition, currentLocation, customerRating, decorStyle, dimensions (width, depth, height)
             era, filling, finish, foldable, hasStorage, intendedRoom, isAdjustable, isAntique, isVintage, madeIn, material, name, ordinal, owner, pattern, price, sequential , soldBy, swivels, upholstery, weight, weightCapacity \\
        \underline{HOLIDAY}&
            duration, endTime, name, startTime \\
        \underline{JACKET}&
            hemLength, hemStyle, jacketStyle, necklineStyle, sleeveLength, sleeveStyle, waistStyle \\
        \underline{LOCATION}&
            city, continent, country, currentDate, currentTime, region, state \\
        \underline{SITUATION}&
            agent, situationLocation, situationTime, situationType, theme \\
        \underline{SIZE}&
            ageSize, alphabeticSize, numericSize, ordinal, sequential, sizeType \\
        \underline{SKIRT}&
            hemLength, hemStyle, skirtStyle, waistStyle \\
        \underline{SWEATER}&
            necklineStyle, sleeveLength, sleeveStyle, sweaterStyle, waistStyle \\
        \underline{USER }&
            attentionOn, name \\
        \bottomrule[\heavyrulewidth]
        \end{tabular}
    \end{center}
    \caption{List of object attributes in the \simmc ontology}
    \vspace*{\captionReduceBot}
    \label{tab:object_attributes}
\end{table*}

\begin{table*}[t]
    \begin{center}
        \setlength\tabcolsep{5pt}
        \begin{tabular}{p{0.15\textwidth}p{0.75\textwidth}}
        \toprule[\heavyrulewidth]
        \underline{CHECK} & \texttt{check} (STRING) \\
        \underline{COMPARE} & \texttt{comp} (OBJECT) \\
        \underline{COUNT} & \texttt{countFrom} (THING), \texttt{countTo} (THING), \texttt{countUnit} (STRING) \\
        \bottomrule[\heavyrulewidth]
        \end{tabular}
    \end{center}
    \caption{List of activity attributes in the \simmc ontology}
    \vspace*{\captionReduceBot}
    \label{tab:activity_attributes}
\end{table*}

\section{API Call List}
\label{sec:appendix_api_call_list}
\reftab{tab:api_call_list} shows the list of all APIs supported in our \simmc datasets.
\begin{table}[t]
    \begin{center}
        \scalebox{0.8}{
        \setlength\tabcolsep{5pt}
        \begin{tabular}{p{0.60\columnwidth}p{0.50\columnwidth}}
        \toprule[\heavyrulewidth]
        \hfil\textbf{API Name} & \hfil \textbf{Arguments}\\
        \midrule
        \multicolumn{2}{c}{\textbf{\furniture}} \\
        \midrule
        \textit{SearchFurniture:} \newline
        Search items using the item attributes
        &
        Category, color, intended room, material, price range, \etc
        \\
        \cdashlinelr{1-2}
        \textit{SpecifyInfo:} \newline
        Get and specify information (attributes) about 
        an item
        & 
        Material, price range (min--max), customer rating, \etc
        \\
        \cdashlinelr{1-2}
        \textit{FocusOnFurniture:}\newline
        Focus on an item to enlarge (for a better view)
        & 
        Position of argument item on the carousel (left, center, right)\\
        \cdashlinelr{1-2}
        \textit{RotateFurniture:}\newline
        Rotate a focused furniture item in the view
        &
        Rotational directions (left, right, up, down, front, back)
        \\
        \cdashlinelr{1-2}
        \textit{NavigateCarousel:}\newline
        Navigate the carousel to explore search results
        &
        Navigating directions (next and previous)
        \\
        \midrule
        \multicolumn{2}{c}{\textbf{\fashion}} \\
        \midrule
        \textit{SpecifyInfo:}\newline
        Get and specify information (attributes) about 
        an item
        & Brand, price, customer rating, available sizes, colors, \etc\\
        \cdashlinelr{1-2}
        \textit{Search(Database$|$Memory):} 
        Select a relevant image from either the database or memory,
        and specify information 
        & Brand, price, customer rating, available sizes, colors, \etc\\
        \bottomrule[\heavyrulewidth]
        \end{tabular}
        }
    \end{center}
    \caption{
        List of APIs supported in our \simmc datasets with attributes.
        We also include \textit{None} as an action when no API call is required 
        and \textit{AddToCart} to specify adding an item to cart for purchase.
    }
    \vspace*{\captionReduceBot}
    \label{tab:api_call_list}
\end{table}

\section{Implementation Details}
\label{sec:appendix_implementation_details}
All our models are trained using PyTorch \cite{pytorch}.
We consider words (after converting them to lowercase)
that occur at least $5$ times in the training set, to yield model dictionaries of size 
$2619$ and $2032$ for \furniture and \fashion, respectively.
We learn $D_W=256$ dimensional word embeddings for each of these words that are fed into
utterance and history encoder.
All the LSTMs ($2$ layers) and Transformers ($4$ layers, $4$ heads each, with $2048$ internal
state) have a hidden state of size $D_H=256$, in our experiments.
We optimize the objective function using Adam \cite{adam_kingma15} with a learning rate
of $10^{-4}$ and clip the gradients by value to be within $[-1.0, 1.0]$.
The model hyperparameters are selected via early stopping on the development set.

\section{Model Visualizations}
\label{sec:appendix_model_visualization}
\begin{figure}[t]
    \centering
    \includegraphics[width=0.45\columnwidth]{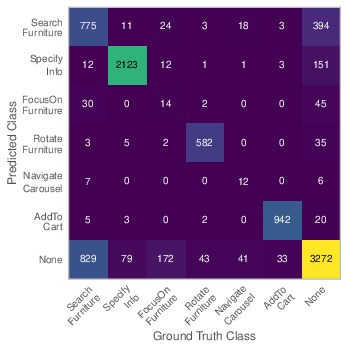}
    \vspace{-1pt}
    \caption{Confusion matrix for hierarchical recurrent encoder (HRE) on \furniture.}
    \vspace{-14pt}
    \label{fig:mn_confusion_matrix_furniture}
\end{figure}

The action API confusion matrix for hierarchical recurrent encoder (HRE) model for the 
\furniture dataset is given in \reffig{fig:mn_confusion_matrix_furniture}.

\section{Dataset Examples}
\label{sec:appendix_examples}

See \reftab{tab:simmc_furniture_ex_1} and \reftab{tab:simmc_fashion_ex_1} in Appendix \ref{sec:appendix_examples} for annotated dialog examples that show our \simmc ontology in action for both our datasets.

\section{Document Changelog}
To help the readers track changes to this document, a brief changelog describing the revisions is provided below:

\textbf{v1}: COLING 2020 anonymity period version. Results on random SIMMC data splits.

\textbf{v2}: COLING 2020 camera-ready version. Results on standard SIMMC data splits.

\begin{table*}[ht]
    \begin{center}
        \scalebox{0.76}{
        \setlength\tabcolsep{5pt}
        \begin{tabular}{p{0.15\textwidth} p{0.35\textwidth} p{0.65\textwidth}}
        \toprule
        \multicolumn{3}{c}{\textbf{Dialog Acts}}\\
        \midrule
        \hfil\textbf{Name}
        & \hfil\textbf{Description}
        & \hfil\textbf{Examples}\\
        \midrule
        \multirow{2}{=}{ASK} 
        &
        \multirow{2}{=}{Used when the main intention of the utterance is
        information seeking, i.e. a question.}
        &
        [DA:ASK:GET:DRESS.price How much is the dress?]\\
        &&
        [DA:ASK:GET:TABLE.color What color is [USER.attentionOn that] table?] \\
        
        \midrule
        \multirow{2}{=}{CONFIRM} 
        &
        \multirow{2}{=}{
        Used when the utterance is asking for or giving confirmation for something that has been said in an earlier turn.}
        & 
        [DA:CONFIRM:GET:DRESS.price One moment while I find the dress's price.]\\
        && 
        [DA:CONFIRM:GET:TABLE.color I'll get that table's exact color information from the catalog.]\\
        
        \midrule
        \multirow{2}{=}{INFORM} 
        & 
        \multirow{2}{=}{
        Used when the main intention of the utterance is information providing.}
        & 
        [DA:INFORM:GET:DRESS.price The dress costs [O.price \$99.99].]\\
        && 
        [DA:INFORM:GET:TABLE.color That table is [O.color hunter green].]\\
        
        \midrule
        \multirow{2}{=}{PROMPT} 
        & 
        \multirow{2}{=}{
        Used when the main intention of the utterance is to suggest an action or prompt the user to take an action.} 
        & 
        [DA:PROMPT:PREFER:DRESS What do you think of the dress?]\\ 
        && 
        [DA:PROMPT:ADD\_TO\_CART:TABLE Would you like me to add the table to your shopping cart?]\\ 
        \midrule
        \multirow{2}{=}{REQUEST} 
        & 
        \multirow{2}{=}{
        Used when the utterance is a request for action.}
        &
        [DA:REQUEST:ADD\_TO\_CART:DRESS I want to buy that dress!] \\
        &&
        [DA:REQUEST:ROTATE:TABLE Show me a [A.rotateTo:SIDE side] view first.]\\ 
        
        \bottomrule
        \toprule
        \multicolumn{3}{c}{\textbf{\Activities}}\\
        \midrule
        \hfil\textbf{Name}
        & \hfil\textbf{Description}
        & \hfil\textbf{Examples}\\
        \midrule        
        \multirow{2}{=}{ADD\_TO\_CART} 
        &
        \multirow{2}{=}{
        Indicates an intent to purchase.}
        &
        [DA:REQUEST:ADD\_TO\_CART:DRESS Add the [O.color green] one to my cart.]\\
        &&
        [DA:INFORM:ADD\_TO\_CART:TABLE I've added the [O.price \$50] table for check out.]\\
        \midrule

        \multirow{2}{=}{CHECK} 
        &
        \multirow{2}{=}{
        Requests a yes/no and alternative questions be answered about an items attribute value.}
        &
        [DA:REQUEST:CHECK:DRESS.color Is the dress [.check green] or [.check blue] ?] \\
        &&
        [DA:INFORM:CHECK:TABLE.color Yes, the table is [.check blue ] .] \\
        \midrule
        
        \multirow{2}{=}{COMPARE} 
        &
        \multirow{2}{=}{
        Requests two (or more) items be compared along a stated attribute.}
        &
        [DA:REQUEST:COMPARE:DRESS.price Is the [R1.color green] [A.comp:DRESS\_1 one] more expensive than [2:USER.attentionOn this] [A.comp:DRESS\_2 dress]?] \\
        &&
        [DA:INFORM:COMPARE:TABLE.width The [R1.color blue] [A.comp:TABLE\_1 table] is wider.] \\
        \midrule
        
        \multirow{2}{=}{COUNT} 
        &
        \multirow{2}{=}{
        Requests the number of items fitting a certain description be returned.}
        &
        [DA:REQUEST:COUNT:DRESS How many [O.color green] ones do you have?] \\
        &&
        [DA:INFORM:COUNT:TABLE I found [A.amount 5 ] [O.color blue] tables.] \\
        \midrule
        
        \multirow{2}{=}{DISPREFER} 
        &
        \multirow{2}{=}{
        Indicates dislike for an item or attribute of that item.}
        &
        [DA:INFORM:DISPREFER:DRESS [USER.attentionOn That] dress is ugly!] \\
        &&
        [DA:INFORM:DISPREFER:TABLE.price I'm not a fan of the cost of the table.] \\
        \midrule
        
        \multirow{2}{=}{GET} 
        &
        \multirow{2}{=}{
        Requests some type of item or attribute of an item be retrieved.}
        &
        [DA:REQUEST:GET:DRESS I'd like to a buy a dress.] \\
        &&
        [DA:INFORM:GET:TABLE.brand This table is made by [O.brand [.name Wind \& Wool]]] \\
        \midrule
        
        \multirow{2}{=}{PREFER} 
        &
        \multirow{2}{=}{
        Indicates like for an item or attribute of that item.}
        &
        [DA:INFORM:PREFER:DRESS [USER.attentionOn That] dress is beautiful!] \\
        &&
        [DA:INFORM:PREFER:TABLE.price Wow what a bargain for the table!]\\
        \midrule
        
        \multirow{2}{=}{REFINE} 
        &
        \multirow{2}{=}{
        Indicates additional constraints to restrict a search.}
        &
        [DA:REQUEST:REFINE:DRESS.color Only show me [O.color green] dresses]  \\
        &&
        [DA:INFORM:REFINE:TABLE.price I've limited results to tables [O.price under \$1000].] \\
        \midrule
        
        \multirow{2}{=}{ROTATE} 
        &
        \multirow{2}{=}{
        Requests an item (of furniture) be rotated to see another view.}
        &
        [DA:REQUEST:ROTATE:TABLE Can you show me the [A.rotateTo:BACK back] of the table?] \\
        &&
        [DA:CONFIRM:ROTATE:TABLE Yes, I'll provide the [A.rotateTo:BACK back] view momentarily.] \\
        
        \bottomrule
    \end{tabular}
    }
    \end{center}
    \vspace{-8pt}
    \caption{List of \textbf{Dialog Acts} and \textbf{\Activities} used in the \simmc Annotation Ontology
    (\refsec{sec:annotations}) along with examples from both \furniture and Fashion (where
    applicable).
    We use a compositional, linearized, and interpretable annotation ontology that is
    unified for both the user and assistant utterances.}
    \label{tab:simmc_flex_ontology}
\end{table*}

\vfill
\pagebreak


\begin{table*}[ht]
    \begin{center}
        \scalebox{0.92}{
        \setlength\tabcolsep{5pt}
        \begin{tabular}{p{0.28\textwidth}p{0.25\textwidth}p{0.39\textwidth}}
        \toprule[\heavyrulewidth]
        \textbf{Situated Context}  & \textbf{Dialog Utterances} & \textbf{Dialog Annotation} \\
        \midrule
           & U: I am looking for table lamps & [DA:REQUEST:GET:TABLE\_LAMP I am look for table lamps] \\
        \midrule
            \parbox[c]{1em}{
                \includegraphics[width=0.28\textwidth]{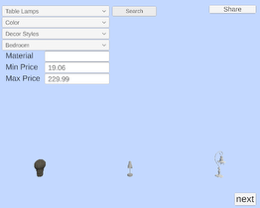}
                }
        & A: Are you looking for bedroom lamps? & [DA:PROMPT:REFINE:TABLE\_LAMP  .intendedRoom Are you looking for [.intendedRoom bedroom] lamps?] \\
        \cmidrule(r){2-3}        
          & U: Yes, bedroom lamps & [DA:INFORM:REFINE:TABLE\_LAMP .intendedRoom Yes, [.intendedRoom bedroom] lamps] \\
        \midrule
          \multirow{2}{*}{
              \parbox[c]{1em}{\includegraphics[width=0.28\textwidth]{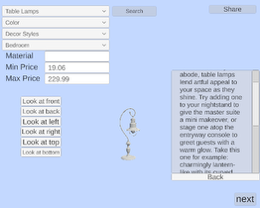}} 
          }
          & A: If you're into a sophisticated style, this neutral gray finish lamp will fit any color palette. I'm showing you the Mentzer Table Lamp. It has a curved steel frame finished in chrome. & [DA:INFORM:GET:TABLE\_LAMP If you're into a [O.decorStyle sophisticated] style, this [O.finish neutral gray] finish lamp will fit any color palette. I'm showing you the [O.name Mentzer] Table Lamp. It has a curved steel frame finished in [O.color chrome].] \\
        \cmidrule(r){2-3}
          & U: What is the cost Mentzer Table Lamp? & [DA:ASK:GET:TABLE\_LAMP .price What is the cost [O.name Mentzer] Table Lamp? \\
        \cmidrule(r){2-3}       
         & A: This is priced at \$127.99 & [DA:INFORM:GET:TABLE\_LAMP .price This is priced at [O.price \$127.99]] \\
        \cmidrule(r){2-3}    
        & U: Do you have anyone under \$50& [DA:REQUEST:GET:TABLE\_LAMP Do you have anyone [O.price under \$50]] \\
        \midrule      
        \multirow{1}{*}{
            \parbox[c]{1em}{\includegraphics[width=0.28\textwidth]{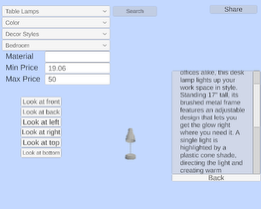}} 
        }
        & A: Yes I do! This Bergh 17 bedroom lamp is priced at \$38. It's 17 inches tall.
        & [DA:CONFIRM:GET:TABLE\_LAMP Yes I do!] [DA:INFORM:GET: TABLE\_LAMP This [O.name Bergh] 17 [O.intendedRoom bedroom] lamp is priced at [O.price \$38]. It's [O.height 17 inches] tall.] \\
        \cmidrule(r){2-3}     
        & U: I’ll take it! & [DA:REQUEST:ADD\_TO\_CART: TABLE\_LAMP I’ll take it!] \\
        \bottomrule[\heavyrulewidth]
        \end{tabular}
        }
    \end{center}
    \caption{\textbf{Dataset Example: SIMMC-Furniture (VR)}}
    \vspace{-10pt}   
    \label{tab:simmc_furniture_ex_1}
\end{table*}

\pagebreak

\begin{table*}[ht]
    \begin{center}
        \scalebox{0.92}{
        \setlength\tabcolsep{5pt}
        \begin{tabular}{p{0.28\textwidth}p{0.25\textwidth}p{0.39\textwidth}}
        \toprule[\heavyrulewidth]
        
        \textbf{Situated Context}  
            & \textbf{Dialog Utterances}
            & \textbf{Dialog Annotation} \\
        \midrule
        
        \multirow{3}{*}{
            \parbox[c]{1em}{
                \includegraphics[width=0.28\textwidth]{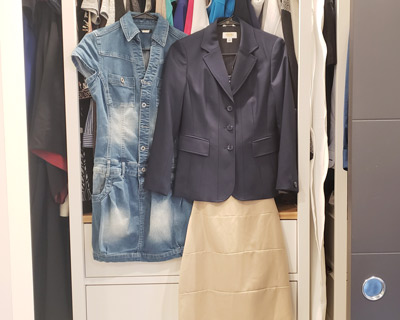}
            }
        }
            & U: What can you tell me about the style of \underline{that brown skirt}?
            & [IN:ASK:GET:SKIRT.skirtStyle What can you tell me about the style of [USER.attentionOn this] skirt?] \\
        \cmidrule(r){2-3}
        
          & A: This style is loose ball gown.
          & [IN:INFORM:GET:SKIRT.skirtStyle This style is [O.skirtStyle loose ball gown]]\\
        \cmidrule(r){2-3}
        
          & U: Can you show me another brown skirt? 
          & [IN:REQUEST:GET:SKIRT Can you show me [O.sequential another] [O.color brown]skirt?] \\
        \midrule
        \multirow{3}{*}{
            \parbox[c]{1em}{
                \includegraphics[width=0.28\textwidth]{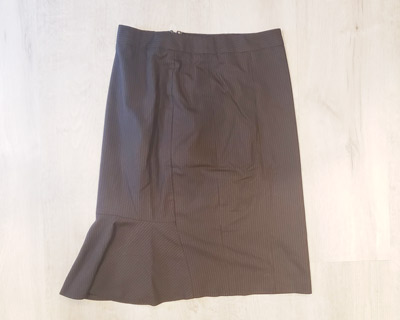}
            } 
        }
         & A: Here's \dashuline{another brown skirt} from Wind \& Wool. 
         & [IN:INFORM:GET:SKIRT Here's [O.sequential another] [O.color brown] skirt from [O.brand [.name Wind \& Wool]].] \\
        \cmidrule(r){2-3}
        
         & U: How much for \dashuline{this one}? 
         & [IN:ASK:GET:SKIRT.price How much for [USER.attentionOn this] one?] \\
        \cmidrule(r){2-3}
        
         & A: \dashuline{This} costs \$139 and has a 3.86 rating. 
         & [IN:INFORM:GET:SKIRT.info This costs [INFO.price \$139] and has a [INFO.customerRating 3.86] rating.] \\
        \midrule
        
        \parbox[c]{1em}{\includegraphics[width=0.28\textwidth]{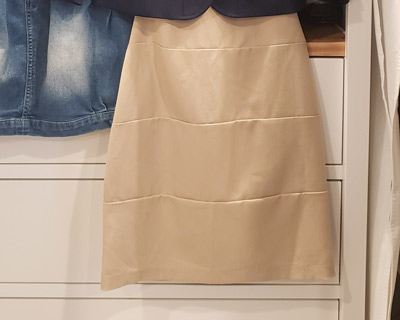}} 
        & U: And how much is \underline{the longer brown} I saw before? 
        & [IN:ASK:GET:SKIRT.price And how much is the [O.hemLength longer] [O.color brown] I saw before?]\\
        \cmidrule(r){2-3}
        
        & A: \underline{That one} is \$272.
        & [IN:INFORM:GET:SKIRT.price That one is [O.price \$272].] \\
        \midrule
        
        \parbox[c]{1em}{\includegraphics[width=0.28\textwidth]{figures/supplementary/img_ex_2}} 
        & U: Put \dashuline{the short brown one} in my cart.
        & [IN:REQUEST:ADD\_TO\_CART:SKIRT Put the [O.hemLength short] [O.color brown] one in my cart.] \\
        \bottomrule[\heavyrulewidth]
        \end{tabular}
        }
    \end{center}
    \caption{\textbf{Dataset Example: SIMMC-Fashion (Image)}. Multimodal coreferences are marked with underlines.}
    \vspace{-10pt}   
    \label{tab:simmc_fashion_ex_1}
\end{table*}

\pagebreak
\end{appendix}

\end{document}